\documentclass[runningheads]{llncs}
\usepackage{graphicx}
\usepackage{booktabs}
\usepackage[ruled,noend]{algorithm2e}
\usepackage{subcaption}
\usepackage{color}
\makeatletter
\newcommand{\printfnsymbol}[1]{%
  \textsuperscript{\@fnsymbol{#1}}%
}
\makeatother

\usepackage{amsmath}
\usepackage{amssymb}
\newcommand{\umang}[1]{{}}
\newcommand{\dimitri}[1]{{}}
\newcommand{\todo}[1]{{}}

\begin{document}
\title{Towards Sparsified Federated Neuroimaging\\Models via Weight Pruning}

\author{Dimitris Stripelis\thanks{equal contribution}\inst{, 1}\and
Umang Gupta\printfnsymbol{1}\inst{, 1}\and Nikhil Dhinagar\inst{2} \and Greg Ver Steeg\inst{1} \and Paul M. Thompson\inst{2} \and Jos\'e Luis Ambite\inst{1}}
\authorrunning{D. Stripelis, U. Gupta et al.}
\institute{Information Sciences Institute, University of Southern California, CA, USA, 90292 \\
\email{\{stripeli, umanggup, gregv, ambite\}@isi.edu}
\and
Imaging Genetics Center, Stevens Neuroimaging and Informatics Institute, University of Southern California, CA, USA, 90292\\
\email{\{dhinagar, thompson\}@ini.usc.edu}}

\maketitle              %
\begin{abstract}
Federated training of large deep neural networks can often be restrictive due to the increasing costs of communicating the updates with increasing model sizes. Various model pruning techniques have been designed in centralized settings to reduce inference times. Combining centralized pruning techniques with federated training seems intuitive for reducing communication costs --- by pruning the model parameters right before the communication step. Moreover, such a progressive model pruning approach during training can also reduce training times/costs. To this end, we propose \textit{FedSparsify}, which performs model pruning during federated training. In our experiments in centralized and federated settings on the brain age prediction task (estimating a person's age from their brain MRI), we demonstrate that models can be pruned up to 95\% sparsity without affecting performance even in challenging federated learning environments with highly heterogeneous data distributions. One surprising benefit of model pruning is improved model privacy. We demonstrate that models with high sparsity are less susceptible to membership inference attacks, a type of privacy attack.

\end{abstract}
\section{Introduction}
Federated Learning~\cite{li2020federated,mcmahan2017communication,yang2019federated} enables distributed training of machine learning and deep learning models across geographically dispersed data silos. In this setting, no data ever leaves its original location, making it appealing for training models over private data that cannot be shared. For these reasons, Federated Learning has witnessed widespread adoption across multiple disciplines, especially in biomedical settings~\cite{dayan2021federated,rieke2020future,sheller2020federated}. Federated training of neural networks involves exchanging/communicating parameters that are updated during local training on private datasets. This parameter exchange incurs high communication costs, limiting the size of neural networks that can be learned~\cite{ro2022scaling}. To circumvent this, model pruning techniques that have been extensively studied in centralized settings~\cite{frankle2018lottery,hoefler2021sparsity,liu2018rethinking} for improving models' training and inference time seem a natural fit towards this direction. 

In this work, we propose a federated training approach incorporating model pruning by directly extending previous work on model pruning in centralized settings~\cite{frankle2018lottery,zhu2017prune}. Similar to these, we use a simple pruning approach of removing weights with the lowest magnitude. However, we consider federated learning environments with heterogeneous data distributions. The learning task is to predict brain age from T1-weighted MRI scans obtained from the UK BioBank dataset~\cite{miller2016multimodal}. We show that with our progressive model pruning strategy, i.e., increasing the sparsity in the model with each federation round, we can learn a neural network model with less than 5\% parameters of the original model while preserving most of the performance.

Even though Federated Learning avoids private data sharing, models trained using federated learning are not always private and may leak sensitive information~\cite{gupta2021membership,pustozerova2020information,zari2021efficient}. This can often be attributed to overfitting or memorization~\cite{gupta2021membership,truex2018towards}. Pruning parameters excessively can reduce the memorization capacity of neural networks. Inspired by this intuition, we evaluate the empirical privacy of the obtained sparsified models through membership inference attacks. We observe that pruned models at extreme degrees of sparsification ($> 95$\%) are less susceptible to membership inference attacks while maintaining learning performance. This suggests a triple win for using pruning during federated training --- a) reduced communication costs, b) reduced inference costs due to small sized final models, and c) reduced privacy leakage. 

Existing federated model pruning strategies focus on reducing the required communication cost during training in order to achieve specific levels of model performance~\cite{bibikar2021federated,jiang2022model}. However, in this work we aim to train highly sparsified models of similar performance to the non-sparsified counterparts while at the same time exploring the privacy gains of federated model sparsification against membership inference attacks. To the best of our knowledge, this is the first work that studies the learning performance and privacy properties of model pruning for deep learning models in the federated neuroimaging domain.

\section{Neuroimaging Learning Environments}

An extensive number of machine learning and deep learning approaches have been recently proposed~\cite{wainberg2018deep} with great success~\cite{ezzati2021predictive,zhu2018image} across multiple biomedical imaging tasks, such as image reconstruction, automated segmentation and predictive analytics. In this work, we evaluate such deep learning approaches for the BrainAGE prediction task over a set of challenging neuroimaging environments in centralized and federated settings.

\vspace{5pt}
\noindent\textbf{Brain Age Prediction Task.} Brain age prediction involves creating a machine learning model to predict a person's chronological age from their brain MRI scan, after training the model on large amounts of data from healthy individuals. When this trained model is applied to new scans from patients and healthy controls, the age difference between each individual's true chronological age and that predicted from their MRI scan has been found to be associated with a broad range of neurological and psychiatric disorders, and with mortality~\cite{cole2015prediction,peng2021accurate}. This age prediction task is formulated as a regression task also known as the Brain Age Gap Estimation (BrainAGE). Various efficient deep learning architectures have been recently proposed based on RNNs~\cite{jonsson2019brain,lam2020accurate} and CNNs~\cite{gupta2021isbi,peng2021accurate} with highly accurate brain age estimations. In our work, we use a 3D-CNN model, similar to~\cite{lam2020accurate,stripelis2021scaling} consisting of seven blocks. The first five blocks are composed of a 3x3x3 3D convolutional layer, instance norm, a 2x2x2 max-pool and ReLU activation functions. The sixth block is a 1x1x1 3D convolutional layer followed by an instance norm and ReLU activation. The final block has an average pooling layer, and a 1x1x1 3D convolutional layer. We test the performance of the model on the BrainAGE task over the UK BioBank dataset~\cite{miller2016multimodal}. Out of the 16,356 subjects with neuroimaging in dataset, we selected 10,446 subjects with no neurological pathology and psychiatric diagnosis as defined by the ICD-10 criteria.

\vspace{5pt}
\noindent\textbf{Centralized Environment.} For centralized training, we follow the same setup as~\cite{gupta2021isbi,lam2020accurate}. We consider 10,466 healthy subjects from the UKBB dataset and split them into train, test and validation sets of sizes 7,312, 2,172 and 940 respectively.  

\vspace{5pt}
\noindent\textbf{Federated Learning Environments.} In our federated learning environment, we consider a centralized (star-shaped) topology~\cite{rieke2020future} where a single controller orchestrates the execution of the participating learners. The controller aggregates learners' local models based on the number of training examples each model was trained on and learners train the global model on their local dataset using Vanilla SGD~\cite{mcmahan2017communication}. We refer to this federated training procedure as \textit{FedAvg}~\cite{mcmahan2017communication}.
\begin{figure*}[htpb]
\centering
  \subfloat[Uniform-IID]{
    \includegraphics[scale=0.2]{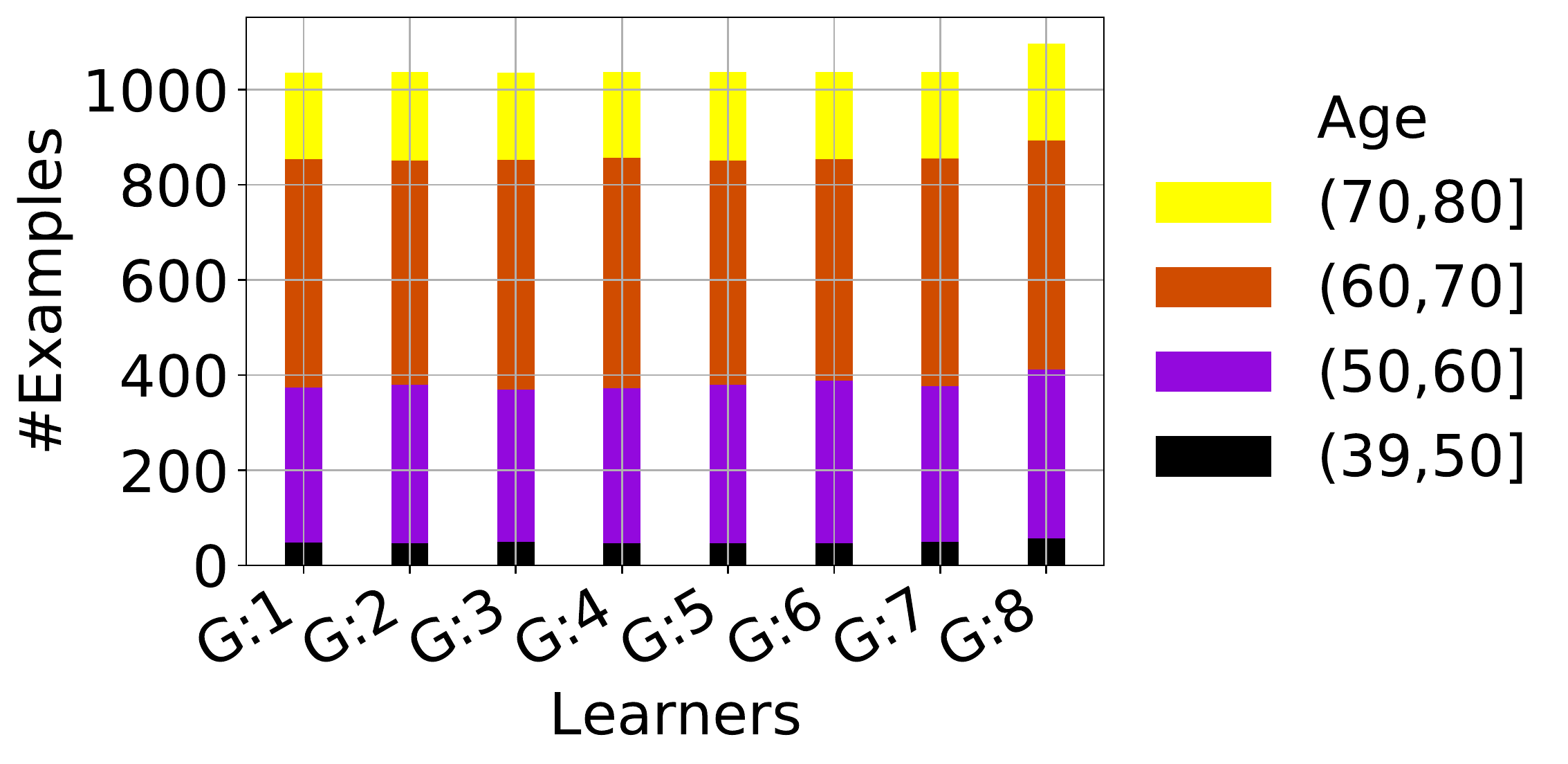}
    \label{subfig:UKBB_AgeBuckets_Uniform_IID}
  }
  \subfloat[Uniform-NonIID]{
    \includegraphics[scale=0.2]{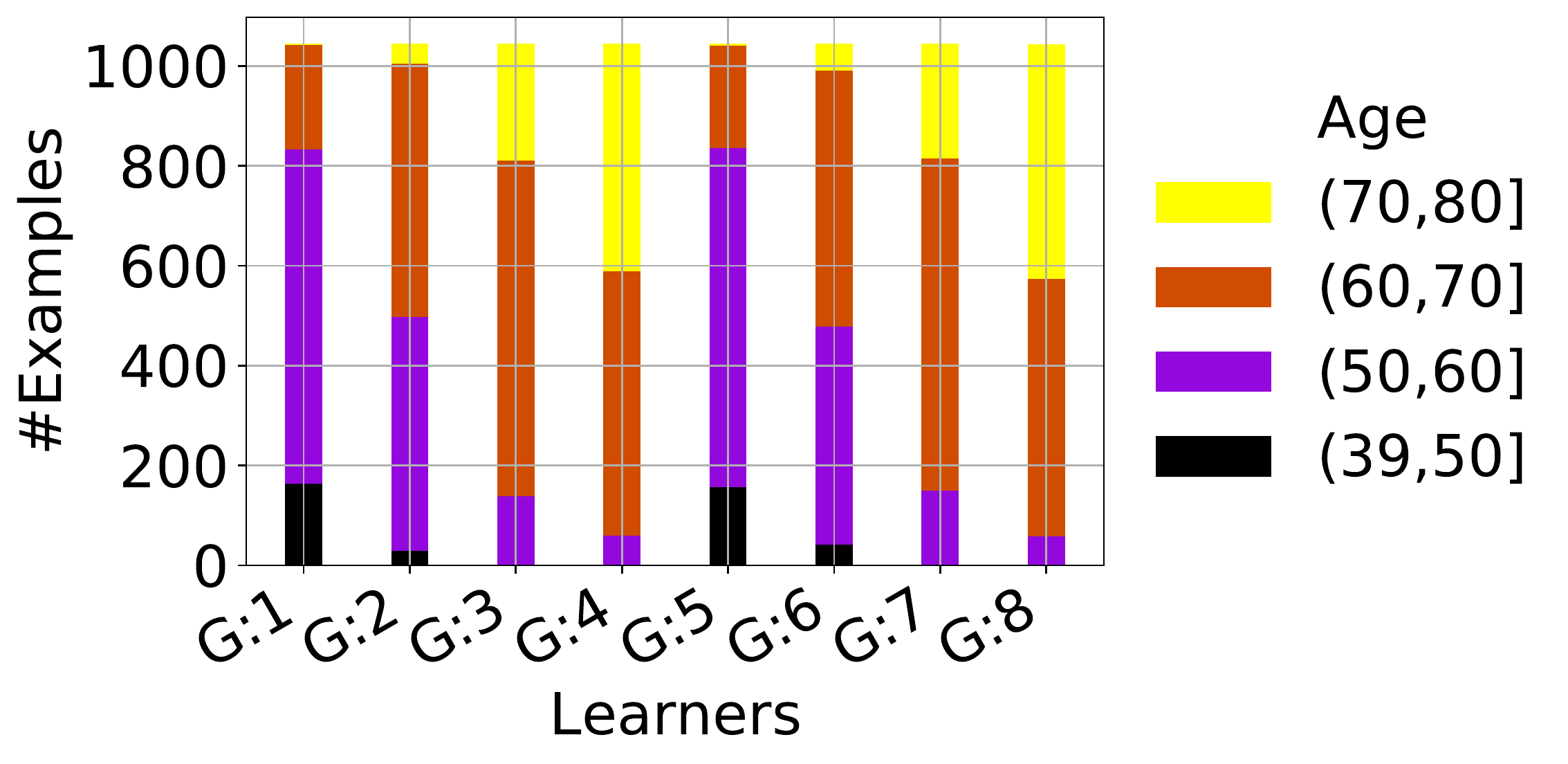}
    \label{subfig:UKBB_AgeBuckets_Uniform_NonIID}
  }
  
    \subfloat[Skewed-IID]{
    \includegraphics[scale=0.2]{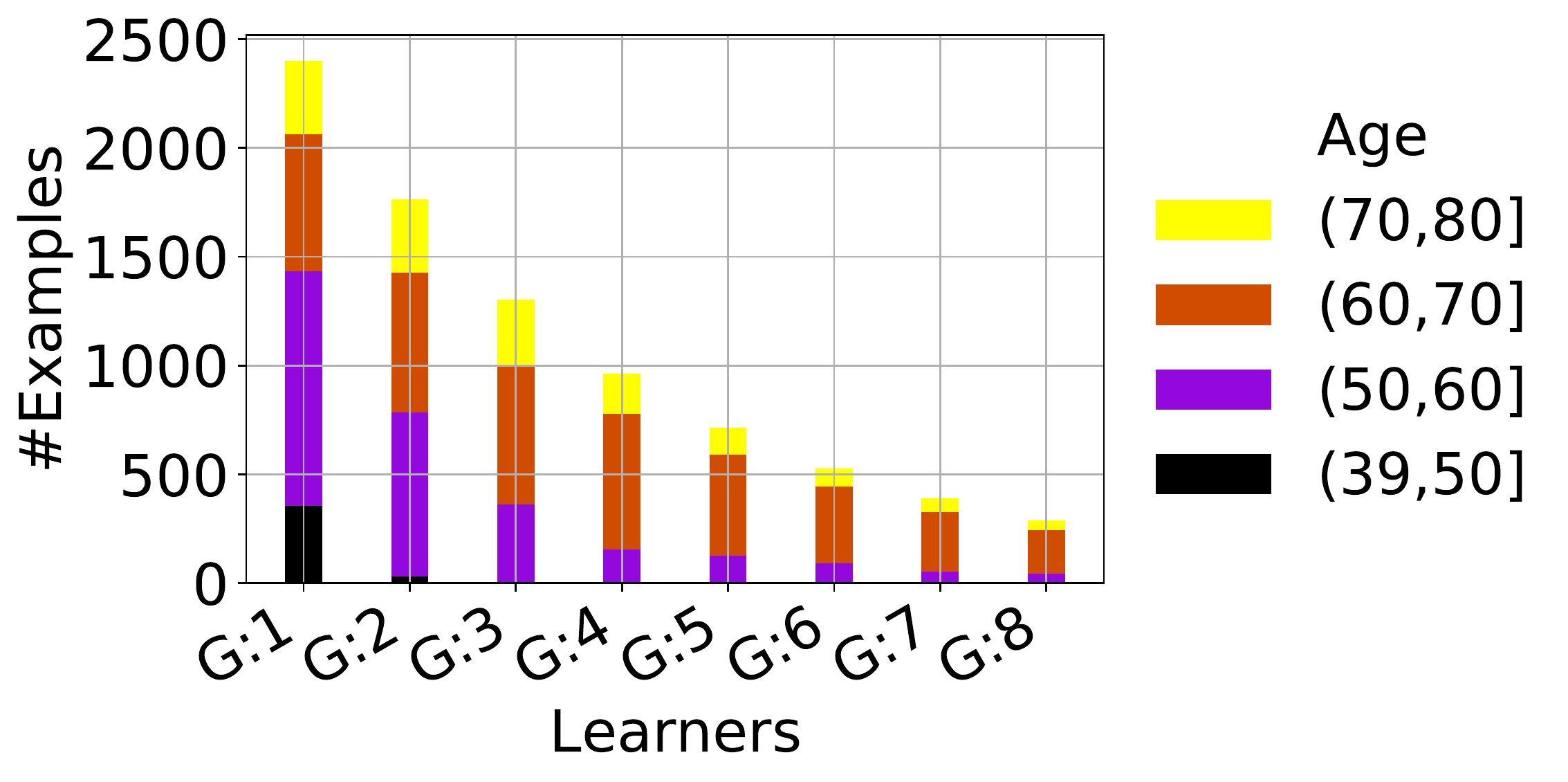}
    \label{subfig:UKBB_AgeBuckets_Skewed_IID}
  }
  \subfloat[Skewed-NonIID]{
    \includegraphics[scale=0.2]{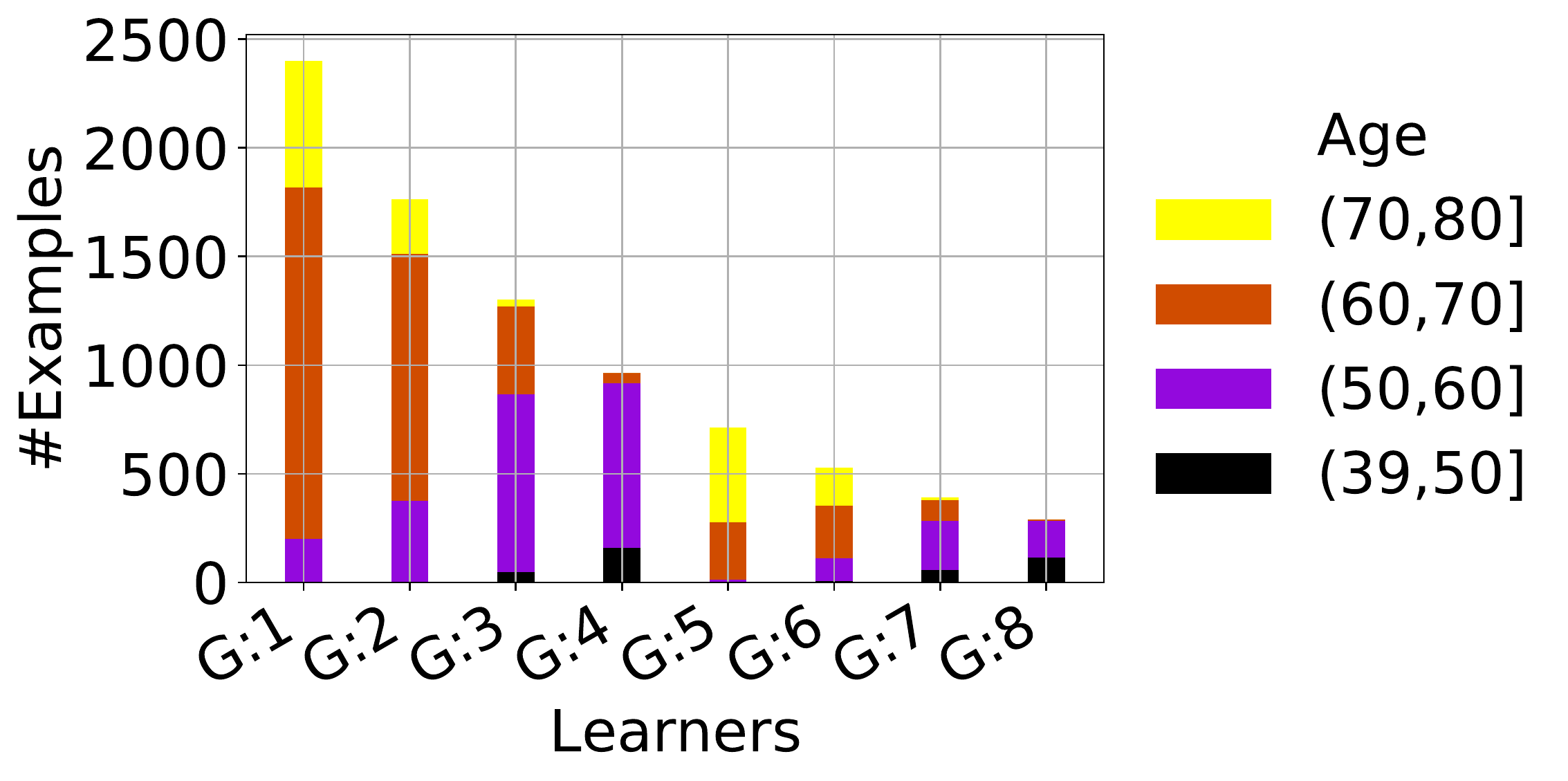}
    \label{subfig:UKBB_AgeBuckets_Skewed_NonIID}
  }

  \caption{UKBB Federated Learning Environments.}
  \label{fig:UKBB_Federation_Age_Distributions}
  
\end{figure*}
Similar to the centralized settings, our learning task is BrainAGE prediction and the learning model is a 3D-CNN~\cite{stripelis2021scaling,peng2021accurate}. We partition the MRI scans of the training and validation datasets from the centralized environment across 8 learners in four federated learning environments~\cite{stripelis2021scaling,stripelis2022semi} of heterogeneous data amounts (Uniform, Skewed) and distributions (IID, Non-IID) per learner (see Figure~\ref{fig:UKBB_Federation_Age_Distributions}). Uniform and Skewed refer to the cases where learners have an equal and rightly skewed number of training samples, respectively. IID and Non-IID refer to the cases where the age range of the local data distribution of the scans owned by a learner captures the global range or a subset, respectively.

\vspace{5pt}
\noindent\textbf{Measuring Privacy via Membership Inference Attacks.} To measure how much information the model leaks about the training set, we consider \textit{Membership Inference Attack}. A {Membership Inference Attack} is often the preferred  approach to evaluate practical privacy leakage from machine learning models~\cite{jayaraman2020revisiting,nasr2019}. Unlike differential privacy which considers worst-case privacy leakage, membership inference attacks can be seen as evaluating average case practical privacy leakages. 
In particular, given a sample (a subject's brain MRI in our case), these attacks infer if the sample was used during training or not. Discovering whether the subject's MRI is in the training set can reveal the personal medical history of the subject, which is undesirable. We use the same attack setups as in~\cite{gupta2021membership}.

In particular, for evaluating models trained in our centralized environment we use their white-box attack setup. We consider access to some actual training and unseen samples for training the attack model; this is a stronger attack setup. One can also launch attacks without accessing actual training samples by training shadow models~\cite{gupta2021membership,nasr2018machine}. We create a balanced test set of training and unseen examples, and report the accuracy of correct predictions as ``attack accuracy''. Lower attack accuracy is more private, and hence better.

For models trained in our federated environments, we consider one of the learners as malicious and launching attacks against other learners. In our federated environments we consider 8 learners, which translates to 56 (7x8) attacks. The learner may train attack models using their private training set and some unseen examples. We report the accuracy of correctly differentiating between other learners' training examples and unseen samples as the ``attack accuracy'' and report the average accuracy, as in~\cite{gupta2021membership}. We also report the number of successful attacks, since due to data heterogeneity not all attacks are successful. We use features derived from the predictions, labels, and gradients of the last two layers of the 3D-CNN to train the attack models.

\section{Model Pruning}
In this section, we discuss model pruning approach for centralized and federated environments for neuroimaging tasks. We evaluate the efficacy of the weight magnitude-based pruning approach on a 3D-CNN trained on centralized and distributed MRI scans.

\vspace{5pt}
\noindent\textbf{Centralized Model Pruning.}
Neural networks can often have redundant parameters which do not affect the outcome. One of the simplest ways of identifying such parameters is by looking at the magnitude of parameters. Parameters with low absolute values do not influence the output much and thus can be safely pruned~\cite{frankle2018lottery,zhu2017prune}. We use this simple approach for pruning. \cite{zhu2017prune} showed that gradual parameters pruning during training is more effective than one-shot pruning at the end. Our federated pruning approach exploits this observation. However, in the centralized setting, we prune in one step at the end of 90\textsuperscript{th} epoch, followed by finetuning for 10 epochs.

\begin{figure}
    \centering
    \includegraphics[width=0.6\linewidth]{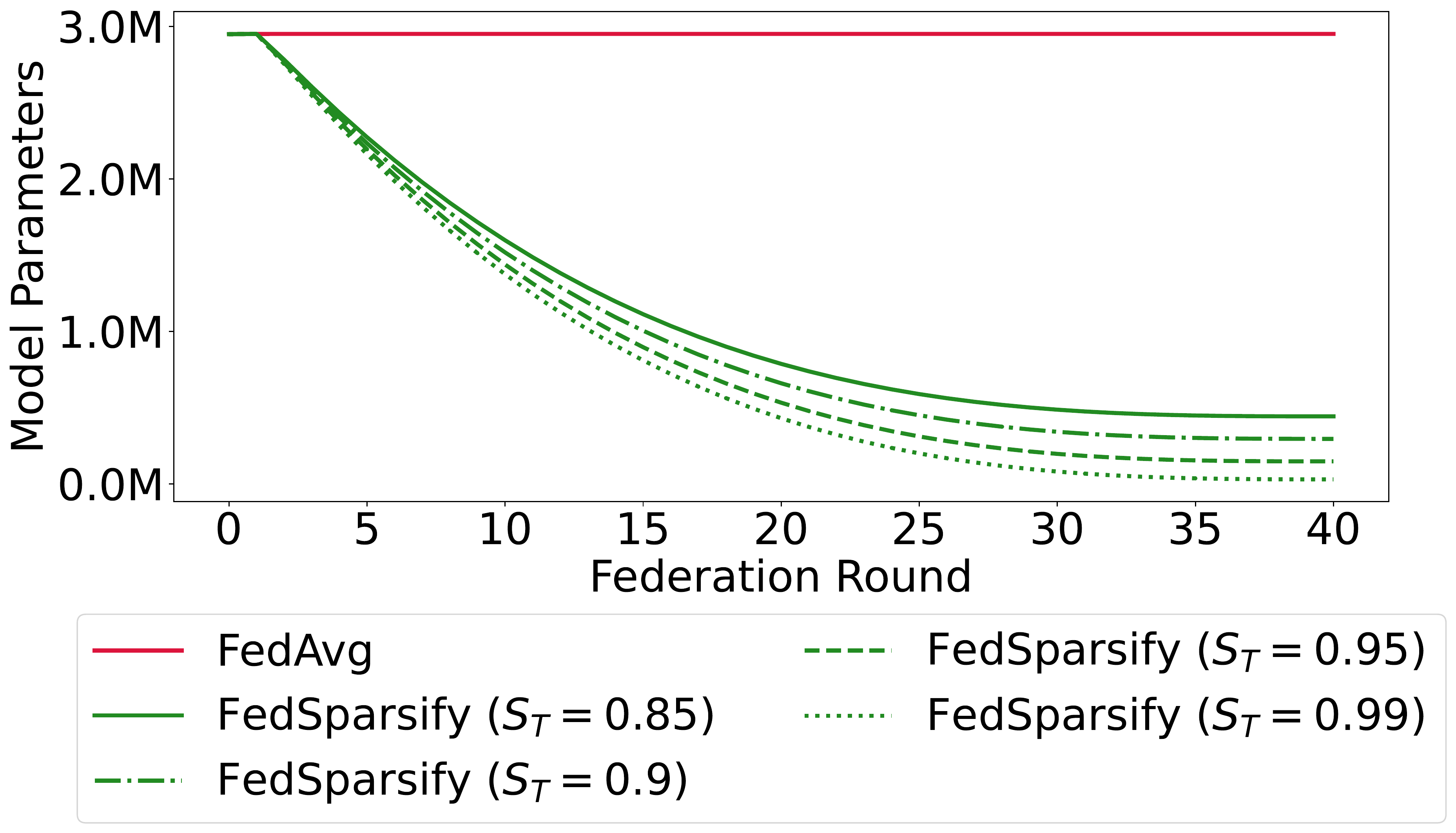}
    \caption{Federated models number of parameters progression with (FedSparsify) and without (FedAvg) sparsification.}
    \label{fig:GlobalModelParametersProgression}
\end{figure}

\noindent\textbf{Federated Model Pruning.} 
We develop our sparsified federated training on top of FedAvg. The global model is pruned at the controller after the controller aggregates the local model updates from the participating learners. Once the new (sparsified) global model is computed, the controller sends  new global model to the learners  along with the associated binary mask representing pruned and unpruned parameters. We use  weight magnitude-based pruning approach~\cite{zhu2017prune} and remove the weights with lowest absolute values. A parameter once pruned is never resurrected. To enforce this during local training, each learner applies the binary mask at every training step (see also Algorithm~\ref{alg:fedpurgemetune} in Appendix).
As we prune during every federation round, our pruning strategy follows a progressive schedule similar to~\cite{stripelis2022federated,zhu2017prune}. The percentage of additional parameters pruned in each round follows an exponentially decreasing schedule, and the overall sparsity at round $t$ is governed by this formula:
\begin{equation}
    s_t = S_T + \left(S_0 - S_T\right)  \left(1 - \frac{F\lfloor t/F\rfloor-t_0}{T-t_0}\right)^ n
    \label{eq:federated_settings_pruning_schedule}
\end{equation}
Here $T$ is total number of federation rounds, $S_0$ and $S_T$ are the initial and desired final sparsity, $F$ is frequency of sparsification, and $t_0$ is the initial sparsification round. The exponent $n$ controls the exponential sparsification rate. We refer to this pruning strategy as \textit{FedSparsify}.
In our experimental evaluation, we explore different final sparsities, i.e., $S_T=$ \{85\%, 90\%, 95\%, 99\%\}. Throughout our experiments, we set the rate of sparsification $n$ to 3, we prune the global model at every federation round, i.e., $F=1$, for a total number of 40 federation rounds, $T=40$, and we start the sparsification schedule at federation round 1, $t_0 = 1$. Figure~\ref{fig:GlobalModelParametersProgression} presents the progression of global model parameters of this sparsification schedule over the course of 40 federation rounds.

\section{Results}
We train the 3D-CNN model\footnote{\url{https://github.com/dstripelis/FedSparsify}} for the brain age prediction task in different learning setups. We perform one-shot pruning in the centralized setup to achieve different sparsity levels. For the federated learning setup, we vary $S_T$, the final sparsity level in Eq.~\ref{eq:federated_settings_pruning_schedule} and prune progressively before communicating updated weights to the learners (see Alg.~\ref{alg:fedpurgemetune}). In all environments the model is trained using Vanilla SGD with a batch size of 1 and learning rate of $1e^{-5}$. During federated training learners train the global model locally for 4 epochs in between federation rounds. All experiments were run on a dedicated GPU server equipped with 4~Quadro RTX 6000/8000 graphics cards of 50~GB RAM each, 31~Intel(R) Xeon(R) Gold 5217 CPU @ 3.00GHz, and 251GB DDR4 RAM.

\begin{figure}
\centering
    \begin{subfigure}{0.48\textwidth}
        \centering
        \includegraphics[width=0.8\linewidth]{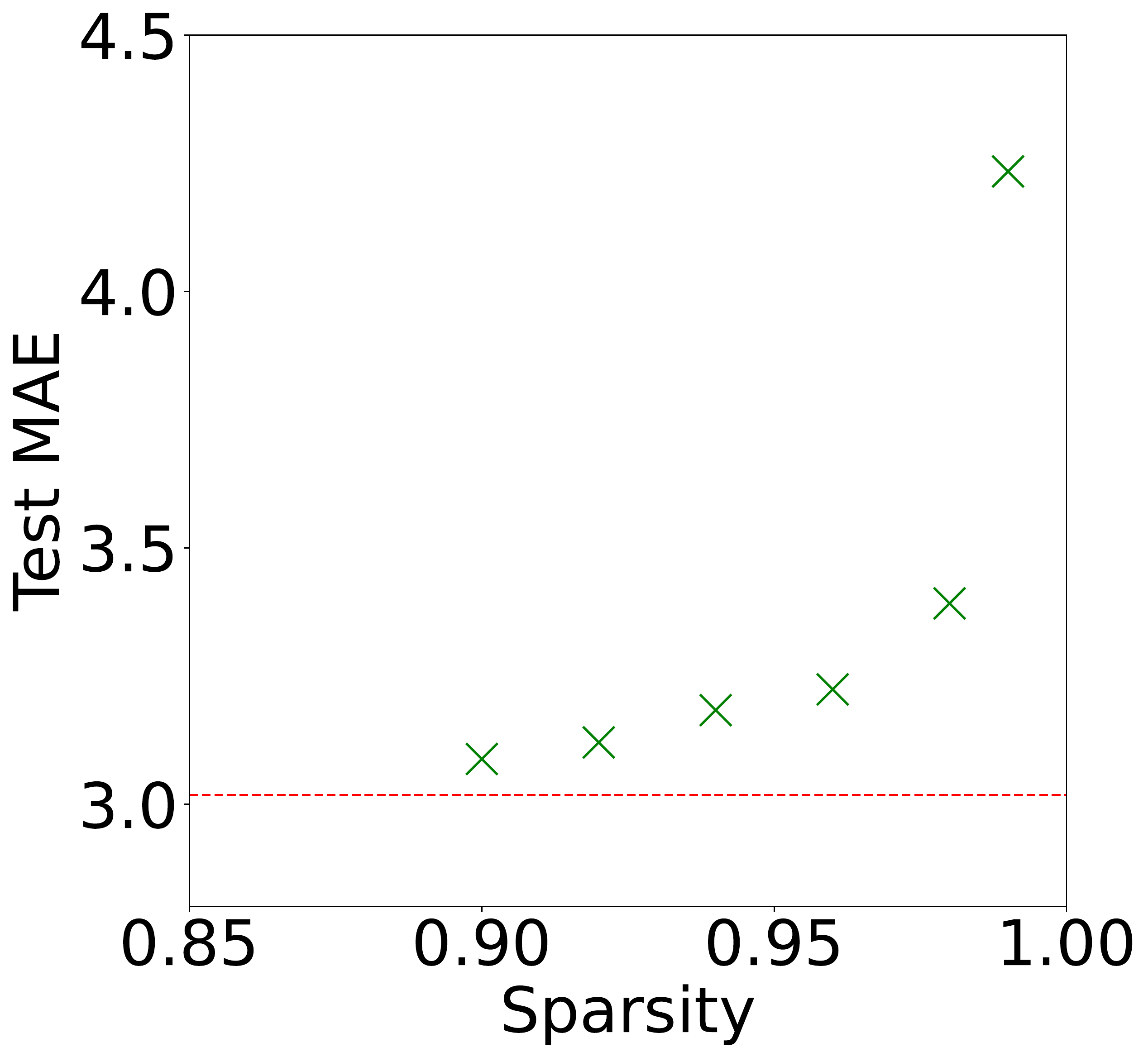}
        \caption{}
        \label{fig:mae_sparsity_tradeoff_centralized}
    \end{subfigure}
    \quad
    \begin{subfigure}{0.48\textwidth}
    \centering
    {
        \includegraphics[width=0.75\linewidth]{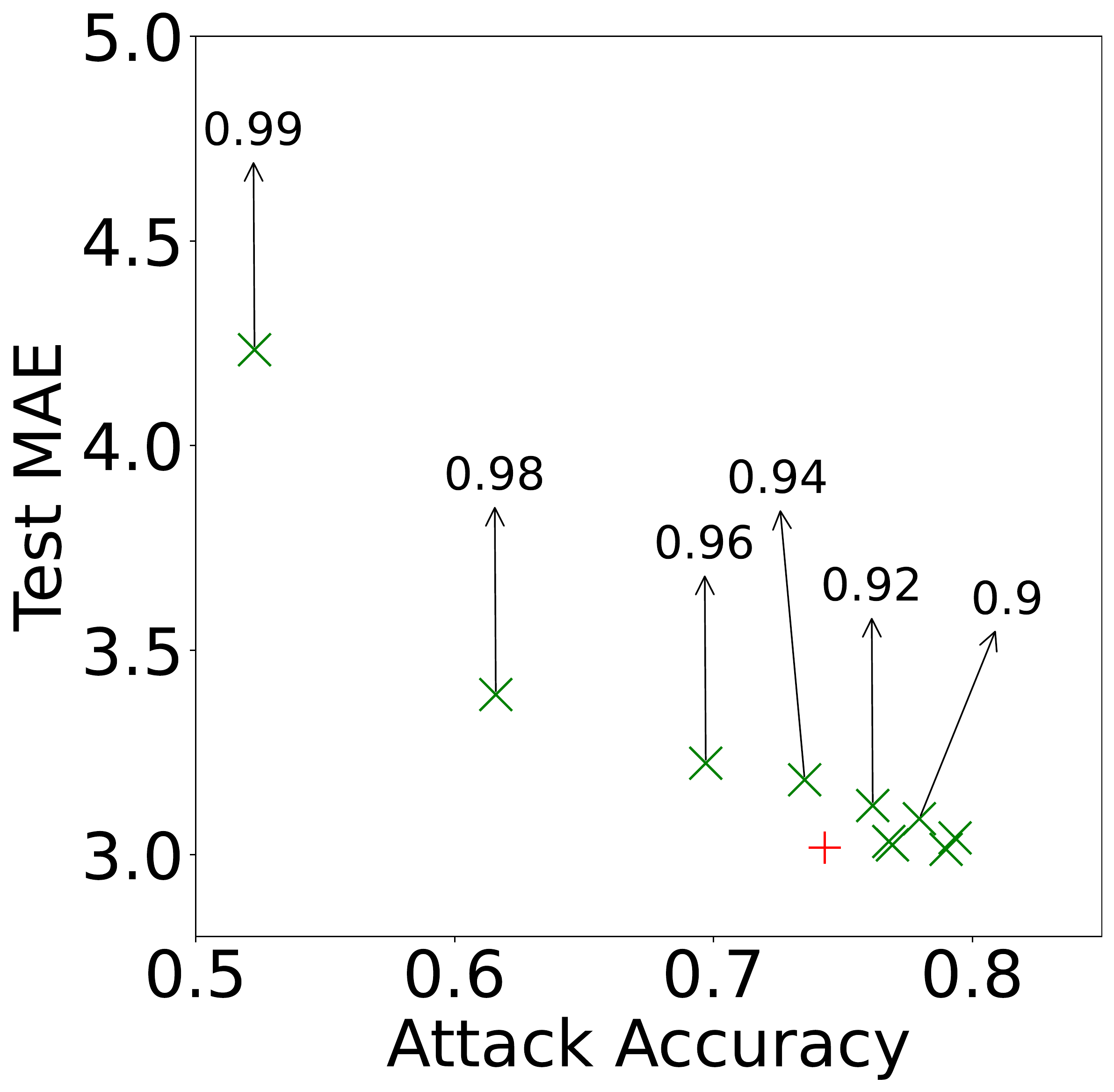}
        \caption{}
        \label{fig:attack_centralized}
    }
    \end{subfigure}
    \caption{Centralized BrainAGE model performance at different sparsity levels (left plot) and model vulnerability to membership inference attacks with respect to model performance (right plot).}
    \label{fig:Centralized_training}
\end{figure}

\begin{figure}
    \centering
    \includegraphics[scale=0.19]{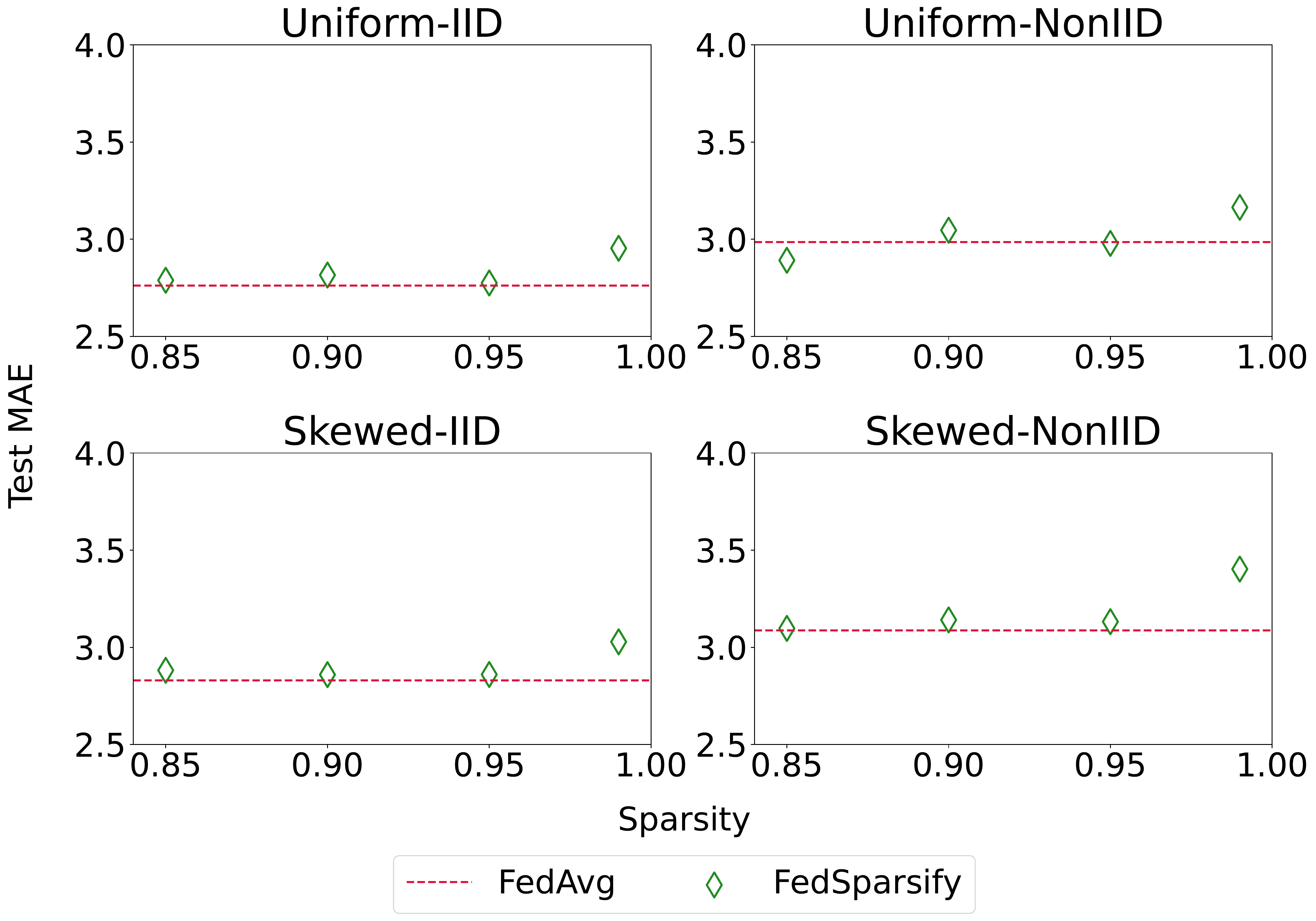}
    \caption{Federated BrainAGE models learning performance at different degrees of sparsification across all four federated learning environments. Dashed line represents performance of non-sparsified model.}
    \label{fig:FederatedModels_Sparsity_vs_LearningPerformance}
\end{figure}

\begin{figure}
    \centering
    \includegraphics[scale=0.19]{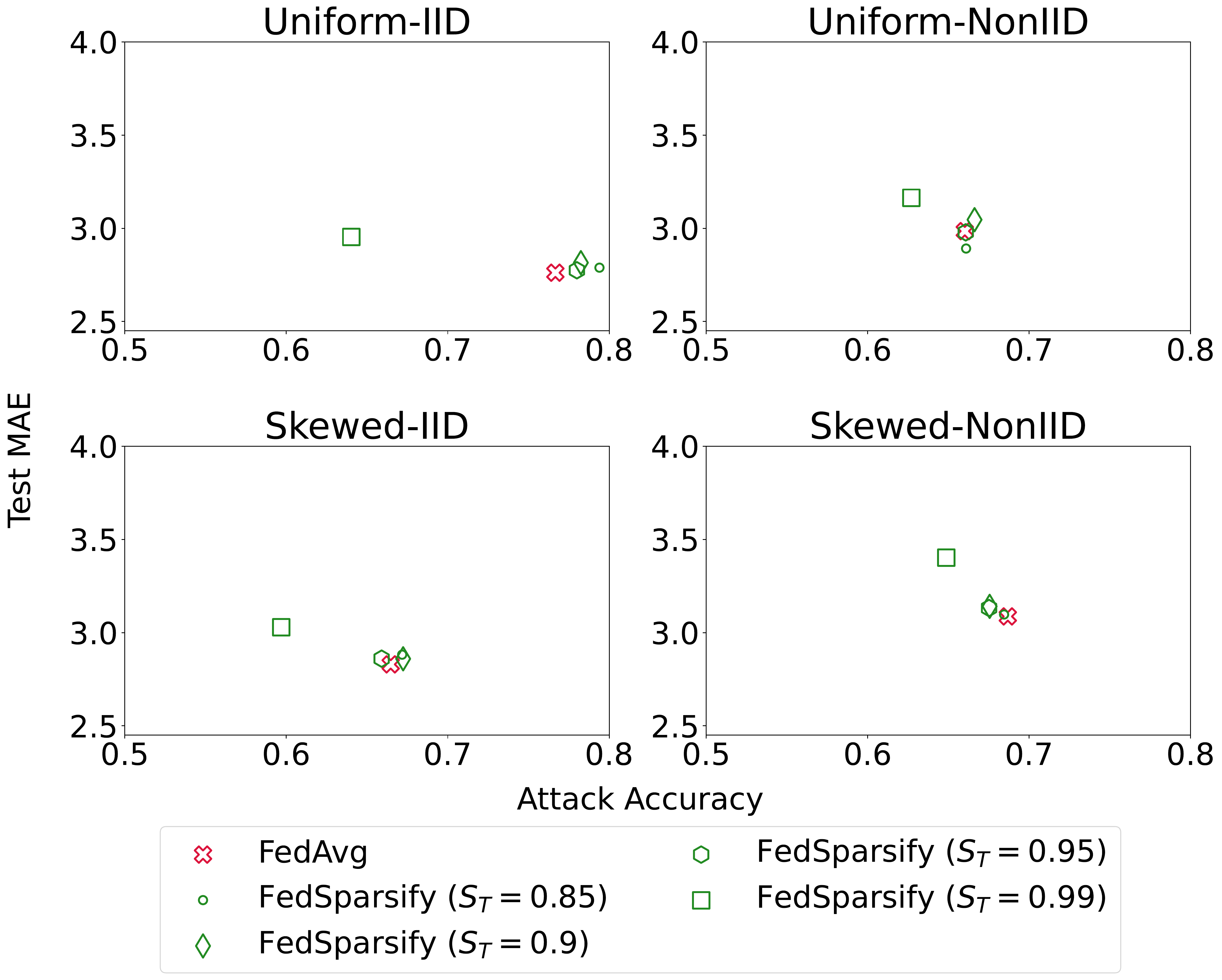}
    \caption{Federated BrainAGE models vulnerability to membership inference attacks with respect to learning performance across all federated environments.}
    \label{fig:FederatedModels_Attack_vs_LearningPerformance}
\end{figure}

\vspace{-20pt}
\begin{table}
    \centering
    \setlength\tabcolsep{3pt}
    \begin{tabular}{@{}cccccccc@{}}
        \toprule
        Sparsity & Params & Size(MBs) & Comm.(MM) & Test MAE & MIA(Success) & Throughput \\ \midrule
        0.0 & 2,950,401 & 10.85 & 1888 & 2.879 & 0.66 (50) & 64.31 \\
        0.85 & 442,561 & 2.09 & 714 & 2.881 & 0.671 (52) & 69.06 \\
        0.9 & 295,041 & 1.43 & 645 & 2.859 & 0.672 (51) & 71.28 \\
        0.95 & 147,521 & 0.73 & 576 & 2.861 & 0.659 (54) & 78.27 \\
        0.99 & 29,505 & 0.16 & 521 & 3.024 & 0.596 (47) & 128.55 \\ \bottomrule
    \end{tabular}
    \caption{Federated Models Comparison in the \textit{Skewed-IID} Environment.}
    \label{tbl:ModelsComparison}
\end{table}

\vspace{5pt}
\noindent\textbf{Model pruning does not hurt performance.}
We first study model performance at different sparsity levels by evaluating the models on a held-out test set. These results are summarized in Fig.~\ref{fig:mae_sparsity_tradeoff_centralized} for centralized training. Even through the one-step pruning approach, we observe that most of the model performance is preserved when 90\% of the parameters are removed. This validates the applicability of weight magnitude-based pruning for deep learning models on neuroimaging tasks. 
We apply our proposed progressive pruning procedure for federated training at different final sparsity levels across four different environments. The results are summarized in Fig.~\ref{fig:FederatedModels_Sparsity_vs_LearningPerformance}. In all cases, model performance is not affected at 95\% sparsity level and performs the same as the {FedAvg} model, which is trained without pruning. Even when only 1\% of the parameters are preserved, i.e., 99\% sparsity, the model performance degrades slightly. 
Table~\ref{tbl:ModelsComparison} provides a quantitative comparison of the total number of parameters and memory/disk size of the final model, the cumulative communication cost in terms of the total number of parameters exchanged during training~\footnote{Communication cost is computed as $\sum_t^T 2 N_Z^t  L $. $T$ represents the total number of federation rounds, $N_Z^t$ the non-zero model parameters at round $t$ and $L$ the number of participating learners. Factor 2 accounts for the model parameters sent from the controller to the learners and from the learners to the controller within a round.}, and the model's learning performance. Our pruning schedule can learn a highly sparsified federated learning model with 3 to 3.5 times lower communication cost than its unpruned counterpart (cf. 521 million to 1888 million parameters). Moreover, the reduced number of the final model parameters also leads to reduced model space/memory footprint, with the sparsified models at 95\% and 99\% sparsification being 67 times smaller than the original model.
Following previous work~\cite{kurtz2020inducing} on model efficiency evaluation~\footnote{\url{https://github.com/neuralmagic/deepsparse}}, we benchmark the inference time for sparse and non-sparse models by recording the total number of processing items per second (i.e., Throughput - items/sec) that each model can perform. Specifically, we take the final model learned with (FedSparsify) and without sparsfication (FedAvg) and stress test its inference time by allocating a total execution time of 60 seconds with a warmup period of 10 seconds. As we show in Table~\ref{tbl:ModelsComparison}, as sparsification increases model throughput increases too, leading to improved inference efficiency especially at 99\% sparsity.

\vspace{5pt}
\noindent\textbf{Excessive Model Pruning may reduce  privacy vulnerability}.
Intuitively, pruning can reduce the ability of a neural network to memorize training data and thus reduce privacy vulnerability. To this end, we evaluate pruned models for privacy leakage using membership inference attacks (Fig.~\ref{fig:attack_centralized} and Fig.~\ref{fig:FederatedModels_Attack_vs_LearningPerformance}). We find that at  extreme sparsity levels ($>$ 95\% for centralized settings and 99\% for federated setting) the attack accuracy reduces suggesting that these models are less vulnerable to privacy leakage compared to non-sparsified models. Compared to the non-sparsified model, the sparsified models are 10\% to 20\% less vulnerable in case Skewed IID and Uniform IID environments, respectively, and 5\% for the Non-IID environments.

\section{Discussion}
We investigated model pruning for deep learning models in the neuroimaging domain through the BrainAGE prediction task in both centralized and federated learning environments. We demonstrated that sparsified models are equally performant as their non-sparsified counterparts even at extreme sparsity levels across all investigated environments. We also evaluated the effectiveness of sparsified models in improving model resiliency against membership inference attacks. We discovered that highly sparsified models could reduce vulnerability to this privacy attack. The vulnerability to membership inference attack is related to the mutual information between the training dataset and activations~\cite{jha2020extension} or model parameters~\cite{farokhi2020modelling}. These results could provide a plausible theoretical explanation as to why pruning reduces the information about the training dataset in neural network weights compared to weights obtained by training without pruning. In the future, we plan to analyze the relation between model sparsification and model privacy and provide a theoretical framework to understand the connection between them better. We also plan to improve model privacy by introducing notions of stochasticity while applying model weight pruning.

\bibliographystyle{splncs04}
\bibliography{refs}

\appendix
\begin{algorithm}
 \caption{\texttt{FedSparsify.} Global model $w$ and global mask $m$ are computed from $N$ participating learners, each indexed by $k$, in current round $t$ out of a total number of $T$ rounds; $E$ is the local training epochs; $s_t$ is the sparsification percentage of model weights at round $t$; $\mathcal{B}$ is the total number of batches per epoch; $\eta$ is the learning rate. $g_k^{(i)}$ denotes gradient of $k$\textsuperscript{th} client's objective with parameters $w_k^{(i)}$. If no sparsification is used the \textit{FedSparsify} algorithm is equivalent to FedAvg.}
    \label{alg:fedpurgemetune}

    \DontPrintSemicolon
    \SetKwProg{Fn}{Procedure}{:}{end}
    \Fn{Controller($w^{(1)}, m^{(1)}$)}{
        \For{$t = 1$ \KwTo $T$}{
            \uIf{\textbf{FedSparsify}}{
                \For{$k = 1$ \KwTo $N$}{
                    $w_k^{(t)} = Client(w^{(t)}, m^{(t)}, E, null)$\\
                }
                $w^{(t+1)} =  \sum_{k=1}^N\frac {|\mathcal D_k|} {|\mathcal D|} w_k^{(t)}$\\
                $m^{(t+1)} = purging\_mask(w^{(t+1)}, s_t)$\\
                ${w}^{(t+1)} = {w}^{(t+1)}\odot m^{(t+1)}$\\
            }
        }
        \textbf{return} $w^{(t+1)}$
    }
    \;
    \Fn{Learner($w, m, E, s_t$)}{
    $w_k^{(0)} = w$\\
    $S = E*\mathcal B$\\
    \For{$i = 0 \ \KwTo\ S$}{
            $w_k^{(i+1)} = w_k^{(i)} -\eta g_k^{(i)}\odot m$
        }
        \textbf{return} $w_k^{(S)}$
    }

\end{algorithm}

\end{document}